%% file: FAME.tex
\documentclass[letterpaper]{article} % DO NOT CHANGE THIS
\usepackage{aaai2026}  % DO NOT CHANGE THIS
\usepackage{times}  % DO NOT CHANGE THIS
\usepackage{helvet}  % DO NOT CHANGE THIS
\usepackage{courier}  % DO NOT CHANGE THIS
\usepackage[hyphens]{url}  % DO NOT CHANGE THIS
\usepackage{graphicx} % DO NOT CHANGE THIS
\usepackage{natbib}  % DO NOT CHANGE THIS AND DO NOT ADD ANY OPTIONS TO IT
\usepackage{caption} % DO NOT CHANGE THIS AND DO NOT ADD ANY OPTIONS TO IT
\usepackage{algorithm}
\usepackage{algorithmic}

\usepackage{newfloat}
\usepackage{listings}
\DeclareCaptionStyle{ruled}{labelfont=normalfont,labelsep=colon,strut=off} % DO NOT CHANGE THIS
\floatstyle{ruled}
\newfloat{listing}{tb}{lst}{}
\floatname{listing}{Listing}
\title{FAME: Fairness-aware Attention-modulated Video Editing}
\author{
    Zhangkai Wu\textsuperscript{\rm 1},
    Xuhui Fan\textsuperscript{\rm 1},
    Zhongyuan Xie\textsuperscript{\rm 1},
    Kaize Shi\textsuperscript{\rm 2},
    Zhidong Li\textsuperscript{\rm 3},
    Longbing Cao\textsuperscript{\rm 1}
}
\affiliations{
    \textsuperscript{\rm 1} Macquarie University, Australia\\
    \textsuperscript{\rm 2} University of Southern Queensland, Australia\\
    \textsuperscript{\rm 3} University of Technology Sydney, Australia
}

\usepackage{bibentry}
\usepackage{booktabs}   
\usepackage{graphicx}    
\usepackage{multirow}    
\usepackage{adjustbox}
\usepackage{amsmath}
\usepackage{amssymb}
\usepackage{cleveref}
\input{math.tex}
\newcommand{\gfun}{fair attribute embedding function}
\newcommand{\ffun}{self-attention manipulation function}
\newcommand{\hfun}{cross-attention reweighting function}
\newcommand{\prom}{soft debiasing prompt encoding}
\newcommand{\selfatt}{temporal self-attention modulation}
\newcommand{\crossatt}{cross-attention reweighting}
\usepackage[table]{xcolor}
\usepackage{multirow}
\usepackage{booktabs}

\begin{document}

\maketitle

\begin{abstract}
Training-free video editing (VE) models tend to fall back on gender stereotypes when rendering profession-related prompts. We  propose \textbf{FAME} for \textit{Fairness-aware Attention-modulated Video Editing} that mitigates profession-related gender biases while preserving prompt alignment and temporal consistency for coherent VE. We derive fairness embeddings from existing minority representations by softly injecting debiasing tokens into the text encoder. Simultaneously, FAME integrates fairness modulation into both temporal self attention and prompt-to-region cross attention to mitigate the motion corruption and temporal inconsistency caused by directly introducing fairness cues. For temporal self attention, FAME introduces a region constrained attention mask combined with time decay weighting, which enhances intra-region coherence while suppressing irrelevant inter-region interactions. For cross attention, it reweights tokens to region matching scores by incorporating fairness sensitive similarity masks derived from debiasing prompt embeddings. Together, these modulations keep fairness-sensitive semantics tied to the right visual regions and prevent temporal drift across frames. Extensive experiments on new VE fairness-oriented benchmark \textit{FairVE} demonstrate that FAME achieves stronger fairness alignment and semantic fidelity, surpassing existing VE baselines. 
\end{abstract}

% \textit{Project page:} {https://anonymous.4open.science/w/FAME-0821/}
\section{Introduction}

Video editing with text prompts has witnessed rapid development thanks to the rise of text-to-image~(T2I) Diffusion Models~(DM) such as Stable Diffusion~(SD)~\cite{rombach2022high} and DALLE-E~\cite{ramesh2021zero}. Building upon these pre-trained T2I models, recent training-free VE methods~\cite{qi_fatezero_2023, geyer2023tokenflow,liu2024video, yang_videograin_2025} enable flexible control over object appearance, motion, and semantics, by manipulating temporal self-attention and cross-frame attention maps. 
% or modifying textual prompts\xuhui{which one? I mean attention maps manipulation or modifying prompt embedding?}. 

% However, while they achieve impressive editing fidelity and efficiency, most works focus solely on spatial–temporal consistency and overlook social considerations such as fairness and bias mitigation.

On the other hand, generating fair content from pre-trained vision-language models has become an increasingly critical problem. \cite{cho2023dall} first revealed that T2I models often reproduce gender and racial stereotypes, even for given neutral prompts~\cite{cho2023dall, shen2023finetuning, gandikota_unified_2024}. This \textit{social bias} largely stems from imbalanced image–text pairs, where data associated with perpetuate majority groups is more frequently collected, and from inherent biases in large-scale language models used in text encoders such as CLIP~\cite{ross2020measuring, birhane2021multimodal}. However, existing fairness-aware T2I methods~\cite{li2024self,kim_rethinking_2025} focus on static images only, which often rely on prompt-image alignment assumptions that may not be generalized to temporally coherent video generation.

\begin{figure}[!htbp]
  \centering
  \includegraphics[trim={5mm 10mm 10mm 10mm}, clip,width=0.9\linewidth]{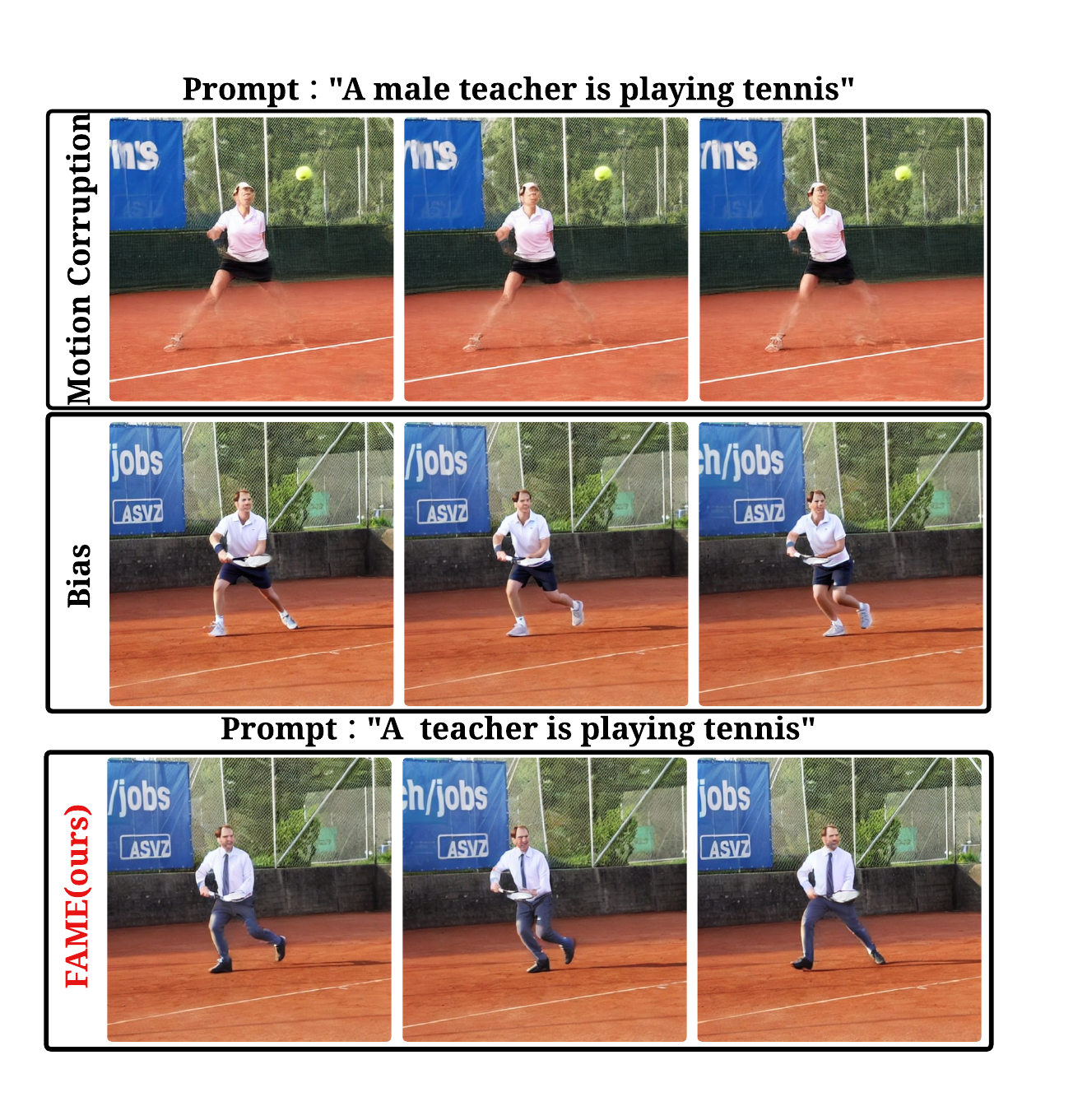}
  \caption{Training-free VE methods (e.g., TokenFlow~\cite{geyer2023tokenflow}) fail to correct gender stereotypes in profession related prompts, as shown in the 1st row, even with debiasing prompt, causing motion corruption, with missing actions, ghosting, and temporal discontinuity. Videograin~\cite{yang_videograin_2025} partly reduces bias but still retains stereotypical features such as outfit, illustrated in the 2nd row. In contrast, FAME combines soft prompt encoding and attention modulation to remove both profession bias and motion corruption, generating coherent videos with accurate debiased profession details.}
  \label{fig:motivation}
\end{figure}

Bridging these two lines of research reveals a significant gap: \textit{fairness-aware VE in a training-free setting has not been formally explored.} Existing T2I models exhibit gender bias toward profession related prompts, defaulting to stereotypical gender content.  When debiasing prompts are applied, these models often suffer from motion corruption, manifesting as semantic mismatch, visual inconsistency, and degraded controllability. More details are revealed in~\Cref{fig:motivation} and our Prompt Responsiveness Test in Appendix \Cref{fig:prompt-test}.These observations highlight the need for a dedicated framework that can achieve fairness control without compromising motion consistency.

To bring fairness into VE while still preserving prompt meaning and motion quality, \textbf{FAME} (Fairness-aware Attention-modulated VE) as a training-free approach tackles both bias and motion corruption together. To reduce gender bias in profession related prompts while preserving their original meaning, we introduce \textit{\prom}, which gently adjusts prompt embeddings toward fairness-aware semantics without altering the intended content. Yet debiasing at the prompt level alone cannot stop fairness cues from fading over time, so we add \textit{\selfatt} to maintain fairness consistency across frames and avoid temporal attention dilution. Finally, since fairness tokens can still drift from their correct visual regions during denoising, we apply \textit{\crossatt} to better align fairness related prompts with the corresponding spatial regions.  

On open source benchmarks, FAME consistently outperforms state-of-the-art training-free VE models in fairness preservation, semantic accuracy, and visual consistency, as measured by both CLIP based metrics and human evaluations. Our main contributions are summarized as follows:
\begin{itemize}
    \item \textbf{FairVE}, as the first benchmark that exploits existing open source datasets and protocols to evaluate fairness in T2I based VE models.
    \item \textbf{FAME}, as a training-free framework to jointly mitigate bias and motion corruption through prompt encoding, temporal attention modulation, and cross attention reweighting at the inference time.
    \item The \textbf{state-of-the-art} performance in fairness preservation, semantic alignment, and temporal coherence by FAME.
\end{itemize}

\begin{figure*}[ht]
  \centering
  \includegraphics[width=0.91\linewidth]{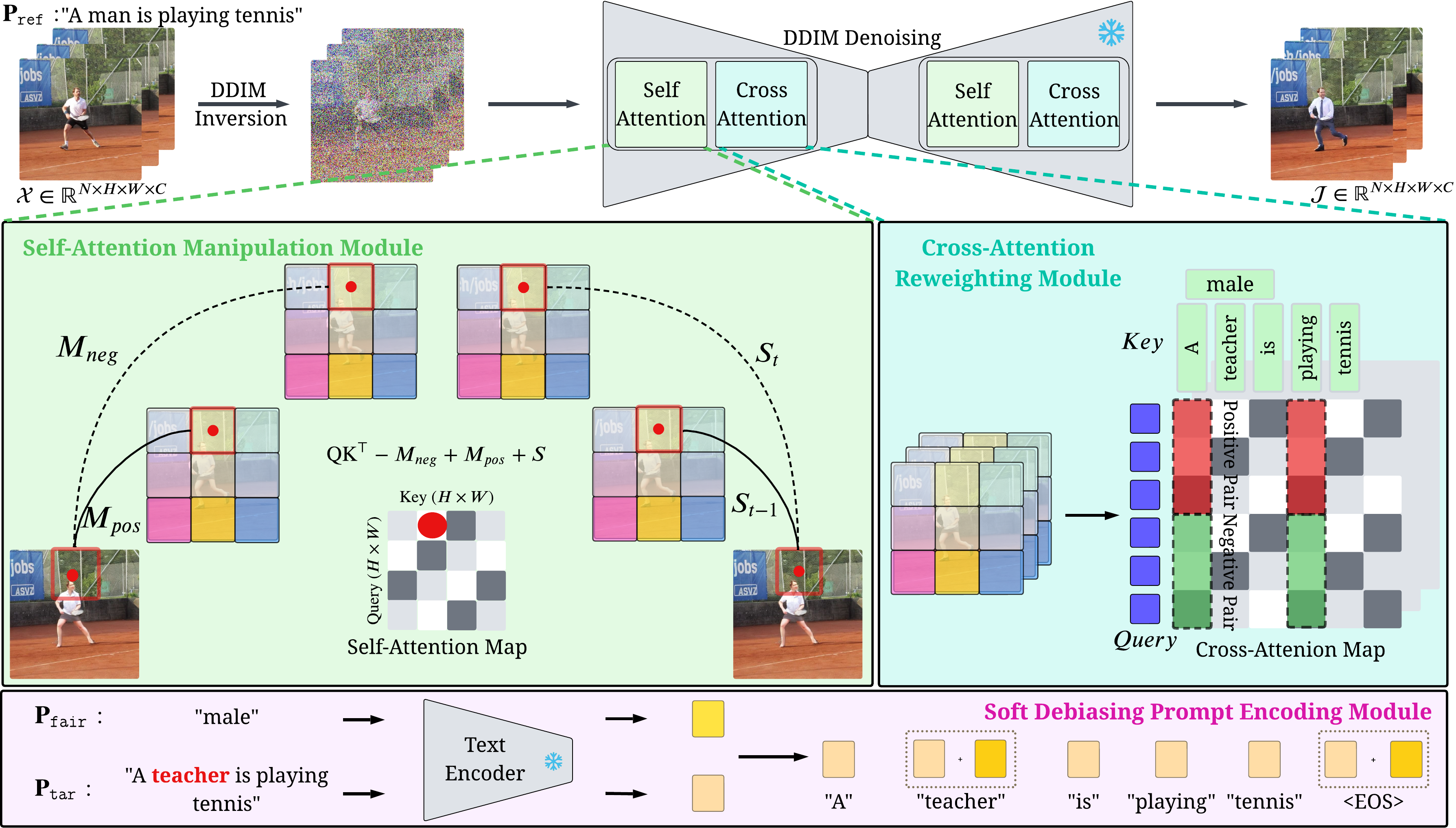}
  \caption{Framework of FAME. In the denoising process, the \prom~module will encode the debiasing prompt $\mbP_{\texttt{tar}}$ explicitly and fuse the self-attention map $\mbQ\mbK^{T}$ with time decay $\mbS$ and element weight $\mbM$ with cross-attention map reweighting.}
  \label{fig:frame}
\end{figure*}

\section{Video Editing Models and Fairness}

\subsection{Video Editing Models}
Latent diffusion models are commonly used, which first encode the input image sequence $\mathcal{X}$ into a latent representation $\mbz_0$ and then gradually corrupt $\mbz_0$ through a forward diffusion process, resulting in intermediate latent variables $\{\mbz_t\}_t$ and their associated attention maps. 
\begin{align*}
q\left(\mbz_t \mid \mbz_{t-1}\right) = &\mathcal{N}\left(\mbz_t ; \sqrt{1 - \beta_t} \mbz_{t-1}, \beta_t \mbI\right), 
% \\
% \mbz_t = &\sqrt{\bar{\alpha}_t} \mbz_0 + \sqrt{1 - \bar{\alpha}_t} \, \mbepsilon,\mbepsilon \sim \cN(\mbzero, \mbI),
\end{align*}
where $t = 1, \dots, T$, with $T$ denoting the total number of steps in the forward diffusion process. The parameter $\beta_t$ controls the noise strength injected at timestep $t$. After defining the forward process, a backward inversion process is used to denoise the noise data into original data, with each step defined as:
\begin{align*}
& p_\mbtheta\left(\mbz_{t-1} \mid \mbz_t\right) = \mathcal{N}\left(\mbz_{t-1} ; \mbmu_\mbtheta(\mbz_t, t), \sigma_t^2 \mbI\right),    
\end{align*}
where $\mbmu_\mbtheta(\mbz_t, t)$ denotes the prediction of $\mbx_0$ at step $t$, and $\sigma_t^2$ is the variance in the inverse step. In practice, the notable U-Net architecture is usually involved in designing the neural architecture of $\mbmu_\mbtheta(\mbz_t, t)$. Further, linear transformed denoising model $\mbepsilon_{\mbtheta}(\mbz_t, t)$ is usually used to predict the residual noise, replacing $\mbmu_\mbtheta(\mbz_t, t)$ for computational convenience as:
\begin{align*}
    \mbz_{t-1} = \sqrt{\frac{\alpha_{t-1}}{\alpha_t}} \mbz_t + \left(\sqrt{\frac{1 - \alpha_{t-1}}{\alpha_{t-1}}} - \sqrt{\frac{1 - \alpha_t}{\alpha_t}}\right) \mbepsilon_{\mbtheta}(\mbz_t, t),
\end{align*}
where $\alpha_t = 1 - \beta_t$ defines the retention ratio of signal. Classifier-free methods~\cite{ho2022classifier} have become a mainstream approach for T2I generation, with the residual noise model $\mbepsilon_{\mbtheta}(\mbz_t, t)$ modified as:
\begin{multline}
 \tilde{\epsilon}_{\mbtheta}\left(\mbz_t, t, \mbphi(\mbp_\texttt{ref}), \emptyset\right) = w \cdot \epsilon_{\mbtheta}\left(\mbz_t, t, \mbphi(\mbp_\texttt{ref})\right) \\
 \quad \quad \quad \quad \quad \quad \quad \quad+(1 - w) \cdot \epsilon_{\mbtheta}\left(\mbz_t, t, \emptyset\right),    
\end{multline}
where $\mbphi(\cdot)$ is the text encoder. The combined noise prediction $\tilde{\epsilon}_{\mbtheta}$ is obtained by linearly interpolating between the conditional prediction (conditioned on the prompt) and the unconditional prediction, with a weighting factor $w \in [0, 1]$ by null-text optimization trick~\cite{ho2022classifier,mokady2023null}. 

% This process is conditioned on the prompt $\mbp_\texttt{ref}$ as:

% \xuhui{, and the cumulative product is given by $\bar{\alpha}_t = \prod_{i=1}^{t} \alpha_i$. The term $\mbz_0$ represents the initial latent code derived from the encoded image sequence $\mathcal{I}$, and $\mbI$ denotes the identity matrix. The random variable $\epsilon$ corresponds to standard Gaussian noise used to perturb $\mbz_0$.}

In training-free VE models, attention maps in the residual noise model $\mbepsilon_{\mbtheta}(\mbz_t, t, \mbphi(\mbp_\texttt{ref}), \emptyset)$ are extracted during the inversion stage to preserve the spatio-temporal consistency across frames. In detail, these models replace the reference prompt $\mbp_{\texttt{ref}}$ with target prompt $\mbp_{\texttt{tar}}$ to predict the noise component $\epsilon_{\mbtheta}(\mbz_t, t, \mbphi(\mbp_\texttt{tar}))$ via the following reverse process:
\begin{multline}
  \tilde{\epsilon}_{\mbtheta}\left(\mbz_t, t, \mbphi(\mbp_\texttt{tar}), \emptyset\right) = w \cdot \epsilon_{\mbtheta}\left(\mbz_t, t, \mbphi(\mbp_\texttt{tar})\right) \\
 +(1 - w) \cdot \epsilon_{\mbtheta}\left(\mbz_t, t, \emptyset\right).
\end{multline}
This guidance mechanism enables controllable generation aligned with textual editing intents while maintaining temporal coherence in video frames.

\subsection{Fairness in Video Editing}
\textbf{T2I-Based VE Models}  
Due to the high inference cost of text-to-video~(T2V) models~\cite{ho2022video,singer2022make,wu2023tune} and the scarcity of high-quality video–text pairs, many recent methods adopt open-source T2I frameworks~\cite{rombach2022high} to enable VE guided by user-provided text or visual cues. 
% The high inference cost of text-to-video~(T2V) models and the lack of high-quality video–text data pairs have led many recent works to adopt opensource T2I frameworks~\cite{rombach2022high} for editing videos verbally or visually on users' intent.
Tune-A-Video~(TAV)~\cite{wu_tune--video_2023} fine-tunes a LoRA module that embeds temporal features to support multi-frame inputs, enabling high-quality and controllable VE. Leveraging the scalability of pre-trained LoRA modules, TAV-based approaches  advance the integration of visual components into the base model. For example, CCEdit~\cite{feng_ccedit_2024} introduces two distinct injection networks to control appearance and structure separately. CAMEL~\cite{zhang_camel_2024} employs CLIP-based prompt embeddings to enhance motion consistency across frames, thereby improving alignment with user intent. DeCo~\cite{leonardis_deco_2025} incorporates a NeRF-based module to improve the 3D perception of human subjects.

\textbf{Training-free T2I-Based VE Models}  
While TAV-based methods incorporate additional trainable components, training-free approaches offer advantages such as faster inference and lower resource consumption. FateZero~\cite{qi_fatezero_2023} introduces cross-frame attention modules to extract spatio-temporal attention maps, guiding multi-frame generation based on prompt variations. Focusing on the training-free pipeline itself, methods such as STEM~\cite{li_video_2024}, WAVE~\cite{leonardis_wave_2025}, SliceEdit~\cite{cohen_slicedit_nodate}, and COVE~\cite{wang_cove_nodate} refine the video inversion stage by selecting key frames to ensure temporal consistency. TokenFlow~\cite{geyer2023tokenflow} and Vid2Me~\cite{li_vidtome_nodate} focus on designing spatio-temporal attention maps to enhance prompt consistency during generation. Another line of work integrates external T2I plugins for fine-grained image to VE frames control. For instance, Ground-A-Video~\cite{jeong_ground--video_2024}, RAVE~\cite{kara_rave_2024}, and VideoShop~\cite{fan_vs_nodate} adopt ControlNet variants, while OCD~\cite{leonardis_object-centric_2025} and VideoGrain~\cite{yang_videograin_2025} leverage the Segment Anything Model to enable object-aware VE. Despite substantial progress in editing quality and task diversity, fairness issues in VE remain largely unaddressed.

\textbf{Fair T2I Generation Models.}~To suppress \textit{social bias} in T2I model generated content, semantic debiasing methods introduce predefined corrective semantic concepts, which are learned either by optimizing the diffusion model in applying classifier-guidance techniques~\cite{brack2023sega,dewan2024diffusion,shen2023finetuning,jung2025multi}, or gradient based modules~\cite{huang2025implicit,azam2025plug,li2025responsible,lee2025localized} to adjust the noise predictor or directly modifying the text prompt embedding space~\cite{kim_rethinking_2025}. However, fine-tuning the entire SD model is computationally expensive, as it requires retraining a large number of parameters. Cross-attention map debiasing approaches~\cite{gandikota_unified_2024, teo2024fairqueue, zhou2024association,jung2024unified,park2025fair} aim to mitigate stereotypical prototypes encoded within the key–query alignment space. In parallel, feature-level debiasing methods~\cite{li2024self,shi2025dissecting} seek to learn fairness-aware embeddings by modulating the intermediate representations within the U-Net bottleneck. Despite notable advances in image generation, fairness concerns in T2I-based VE models remain largely underexplored.

\section{Methodology}
\subsection{Prompt Responsiveness Test for Fairness Verification in Video Editing}
\label{sec:method}
Although various T2I-based debiasing methods~\cite{shen2023finetuning,gandikota_unified_2024,kim_rethinking_2025} have demonstrated that debiasing regions in the latent space of SD variants can be effectively explored, either through classifier-free guidance to embed fair content or by incorporating priors into pre-trained model components, it still remains unclear whether such fairness-guided mechanisms remain effective when extended to VE models. For instance, in recent training-free VE models, cross frame attention mechanisms are introduced to ensure temporal consistency, which fundamentally alters the original prompt semantic coherence, i.e., the alignment between prompt semantics and visual features that underlies T2I models.
 
 As a result, directly transferring fairness control strategies from T2I to VE remains a challenging and open research problem. To assess whether training-free VE models still preserve debiasing regions, we propose the \textit{Prompt Responsiveness Test} designed to evaluate the model's ability to respond to explicitly debising textual conditions.

\textbf{Prompt Responsiveness Test}~To assess whether VE models can preserve fairness-related cues in textual prompts, the prompt responsiveness test can be implemented by incorporating a gender-specific hint with the original professions (e.g., $\mbP_{\texttt{tar}}=$ \texttt{``a male teacher is playing tennis"}) directly into the target editing promt. These gender-profession prompt templates are adapted from prior work~\cite{kim_rethinking_2025}. We then examine whether the generated videos visually reflect the intended gender identity, and whether the direct use of debiasing prompts affects the editing performance.

We apply this test in several training-free VE models. The resulted video (line 2-4 in Appendix~\Cref{fig:prompt-test}) fails to align with the prompt, when gender is explicitly specified in the prompt. This misalignment may be traced back to the introduction of cross-frame attention and latent feature fusion. That is, directly transferring T2I's prompt-based fairness control to VE might not work. An alignment-aware debiasing strategy is required for the fairness VE. More details and visulized results can be referred to the Appendix~\Cref{fig:prompt-test}.

\subsection{The Training-free FAME Model}

% The above Prompt Responsiveness Test shows that existing fair T2I methods rely on static spatial representations, which are incompatible with the cross-frame attention mechanisms in video editing. 
We observe three key challenges when extending fairness-aware image editing methods to VE: (1)~directly injecting debiasing prompts may disrupt the semantic structure of generated content; (2)~fairness-aware local features tend to be diluted by global feature fusion, especially over temporal sequences; and (3)~prompt-to-region attention may become inconsistently distributed across frames. 

Addressing these challenges, we introduce FAME, a training-free VE framework that incorporates fairness into generation through three key components: (1) the \textit{\prom} module, which injects fairness intent into prompts with minimal semantic disruption; (2) the \textit{\selfatt} module, which preserves intra-region coherence across frames; and (3) the \textit{\crossatt} module, which improves alignment between fairness-sensitive tokens and visual regions. All modules are operated at the inference time and no model retraining is required.

\noindent\textbf{Overall Framework.}
Given an input video $\mathcal{X} = \{\mbX^{1}, \mbX^{2}, \dots, \mbX^{N}\}$, where each frame $\mbX^{n} \in \mathbb{R}^{H \times W \times C}$ and the $N$, $H$, $W$, and $C$ denote the frame number, height, width, and channel dimensions, respectively, we consider a fairness-aware VE task guided by a composite textual prompt. This prompt comprises a reference prompt $\mbp_\texttt{ref}$ (e.g., \texttt{``A man is playing tennis."}), a target prompt $\mbp_\texttt{tar}$  (e.g., \texttt{``A teacher is playing tennis."}), and a debiasing attribute indicator $\mbp_{\texttt{fair}}$  (e.g., \texttt{``Male"}), which specifies the direction of bias to be corrected. The objective is to generate an edited video $\mathcal{J} = \{J^1, \dots, J^N\}$, where each frame $J^n \in \mathbb{R}^{H \times W \times C}$ reflects the intent of the edited prompt, while maintaining temporal consistency across frames and alignment with fairness constraints. The whole pipeline is illustrated in~\Cref{fig:frame}.

The proposed FAME is a fairness-aware VE framework that integrates prompt debiasing and attention modulation.  First, we define a \gfun~ $g(\mbp_\texttt{tar}, \mbp_{\texttt{fair}})$ that encodes a selected subset of fairness-sensitive prompt tokens into embeddings for debiasing target text. Each $\mbp_{\texttt{fair}}$ is encoded using an SD text encoder. Second, leveraging the fairness-aware text embedding's attention maps during the inversion stage, we introduce two functions: \ffun~$f(\cdot)$ and \hfun~$h(\cdot)$, to produce a region-wise attention maps $\{\mba_k\}_k\in [0,1]^{H \times W}$. Each entry $a_{k,hw} $ denotes the spatial attention mask value of the $k$-th region relevant to the $w$-th fairness attribute. 

As a result, the process of VE can be described by a model $\mathcal{M}(g,f,h; \mbp_\texttt{ref}, \mbp_{\texttt{tar}}, \mbp_{\texttt{fair}})$, wherein $f(\cdot)$ and $h(\cdot)$ jointly modulate spatio-temporal attention during inference to produce the final edited video $\mathcal{J}$ in alignment with the intended fairness objective.

\subsubsection{Soft Debiasing Prompt Encoding for Fairness Preservation.} Given a fairness attribute prompt $\mbp_{\texttt{fair}}$ (e.g., \texttt{``female"} or \texttt{``male"}), the CLIP-based text encoder $\mbphi(\cdot)$ produces the token embeddings $\mbe_{\texttt{fair}} = \mbphi(\mbp_{\texttt{fair}}) \in \mathbb{R}^{l_1 \times d}$ from $\mbp_{\texttt{fair}}$, where $l_1$ is the number of tokens and $d$ is the embedding dimension. Similarly, the target prompt $\mbp_{\texttt{tar}}$'s token embedding is obtained as $\mbe_{\texttt{tar}} = \mbphi(\mbp_{\texttt{tar}}) \in \mathbb{R}^{l_2 \times d}$. The soft debiasing prompt encoding may be proceeded in two steps without requiring additional training or architectural modifications. 

In the first step, we identify a set of token positions $\mathcal{P} \subseteq \{1, \dots, l_2\}$ in $\mbp_{\texttt{tar}}$ that are semantically relevant to fairness and softly fuse their embeddings with the corresponding fairness tokens:
\begin{equation}
\tilde{\mathbf{e}}_k = \lambda_k \cdot \mbe_{\texttt{tar},k} + (1 - \lambda_k) \cdot \mbe_{\texttt{fair},k}, \quad \text{for } k \in \mathcal{P},
\end{equation}
where $\lambda_k \in [0,1]$ is a mixing coefficient, either fixed (e.g., 0.5) or derived from attention-based importance scores. Tokens outside $\mathcal{P}$ remain unchanged, resulting in an intermediate embedding matrix $\tilde{\mbe}_{\texttt{tar}} \in \mathbb{R}^{l_2 \times d}$.

In the second step, we further refine the debiasing effect by replacing the End of Sequence~(EOS) token embedding in $\tilde{\mbe}_{\texttt{tar}}$ with the corresponding token from $\mbe_{\texttt{fair}}$. This is implemented via a binary injection mask $\mathbf{M} \in \{0,1\}^{l_2 \times d}$ that activates only at the EOS position by \gfun\ $g(\cdot)$:
\begin{equation}
g(\mbp_{\texttt{tar}}, \mbp_{\texttt{fair}}) = \tilde{\mbe}_{\texttt{tar}} + \mathbf{M} \otimes (\mbe_{\texttt{fair}} - \tilde{\mbe}_{\texttt{tar}}),
\end{equation}
where the subtraction and injection are applied only at the EOS token location, since the EOS token in diffusion text encoders aggregates global prompt semantics, allowing fairness information to be injected without disrupting local word-level alignments.

This two-step fusion strategy allows fairness cues to be softly injected into semantically meaningful regions while explicitly guiding the global prompt embedding via EOS-level replacement. Compared to static prompt editing, our method enables more interpretable and localized debiasing, preserving the semantics of original prompt and reducing latent space drift.

\subsubsection{Enhancing Intra-region Coherence in Temporal Self-Attention.}  
To enforce temporal coherence within fairness-sensitive regions, we enhance the temporal self-attention mechanism in the latent space by selectively amplifying intra-region attention while suppressing inter-region interference. Given an input latent tensor $\mbz_{t} \in \mathbb{R}^{h \times w \times \ell \times c}$, we reshape it with a temporal attention preparation function $\varphi$:
\begin{equation}
\varphi: \mathbb{R}^{h \times w \times \ell \times c} \rightarrow \mathbb{R}^{(h \cdot w) \times \ell \times c},
\end{equation}
which flattens the spatial dimensions and aligns features along the temporal axis. The resulting features are projected into query, key, and value matrices:
\begin{equation}
\mbQ = W_\mbQ \cdot \varphi(\mbz_t), \quad
\mbK = W_\mbK \cdot \varphi(\mbz_t), \quad
\mbV = W_\mbV \cdot \varphi(\mbz_t),
\end{equation}
where $W_\mbQ, W_\mbK, W_\mbV \in \mathbb{R}^{c \times d}$ are learnable matrices.

Inspired by self-attention fusion methods~\cite{chefer2023attend,liu2024towards,karmann2025repurposing}, we define the fairness-aware self-attention operator with \ffun\ $f(\cdot)$ as:
\begin{equation}
\texttt{FairAttention}(\mbQ,\mbK,\mbV) = \text{softmax}\left(f(\mbQ, \mbK)\right) \cdot \mbV,
\end{equation}
which incorporates fairness-aware modulation and structural guidance:
\begin{equation}
f(\mbQ, \mbK) = \frac{\mbQ \mbK^\top + \lambda \mbM^{\text{fair}} + \mbmu \mbS}{\sqrt{d}},
\end{equation}
where $\lambda$ and $\mbmu$ are modulation coefficients, and $\mbM^{\text{fair}}$ is defined as
\begin{align}
\mbM^{\text{fair}} = \mbR \odot \mbM^{\text{pos}} - (1 - \mbR) \odot \mbM^{\text{neg}},
\end{align}
with positive and negative maps computed as:
\begin{equation}
\begin{aligned}
\mbM^{\text{pos}} &= \max(\mbQ \mbK^\top) \;-\; \mbQ \mbK^\top, \\
\mbM^{\text{neg}} &= \mbQ \mbK^\top \;-\; \min(\mbQ \mbK^\top),
\end{aligned}
\end{equation}
 respectively, 
and the binary region indicator $\mbR$ is defined element-wise as: \begin{equation}
\mbR[x, y] =
\begin{cases}
1, & \text{if } \texttt{region\_id}(x) = \texttt{region\_id}(y), \\
0, & \text{otherwise}.
\end{cases}
\end{equation}
Here, $\mbM^{\text{fair}}$ modulates attention logits by reinforcing similarities within the same fairness-sensitive region (via $\mbM^{\text{pos}}$) and weakening correlations across different regions (via $\mbM^{\text{neg}}$). The binary mask $\mbR$ simply acts as a selector: it is 1 when two tokens belong to the same region and 0 otherwise, enabling region-specific adjustment.

To further provide region-aware flexibility, we define a soft similarity mask $\mbS \in \mathbb{R}^{(h \cdot w) \times (h \cdot w)}$ based on temporal feature statistics:
\begin{equation}
\mbS[\texttt{q}_1, \texttt{q}_2] = \exp\left( - \frac{\| \mathbf{f}_{\texttt{q}_1} - \mathbf{f}_{\texttt{q}_2} \|^2}{\tau^2} \right),
\end{equation}
where each $\mathbf{f}_{\texttt{q}}$ is the mean feature vector at spatial position $\texttt{q}$ across $\ell$ frames $\mathbf{f}_{\texttt{q}} = \frac{1}{\ell} \sum_{t=1}^{\ell} \mbz_t[\texttt{q}]$, and $\tau$ controls the sharpness of the similarity response. The soft mask $\mbS$ enables continuous region-aware attention modulation, providing a fairness-aligned bias signal complementary to the discrete mask $\mbR$.

\subsubsection{Prompt-to-Region Alignment Across Frames}  
To enhance alignment between fairness-related textual concepts and visual regions across frames, we propose a fairness-aware cross-attention modulation strategy that adjusts attention scores during the denoising process in the latent space.

Specifically, we define a fairness-aware cross-attention operator with \hfun\ $h(\cdot)$:
\begin{equation}
\texttt{FairCrosAttn}(\mbQ_t, \mbK_t, \mbV_t) = \mathrm{softmax}\left(h(\mbQ_t, \mbK_t)\right) \cdot \mbV_t,
\end{equation}
where the attention weight reweighting function $h(\mbQ_t, \mbK_t)$ incorporates fairness-guided modulation:
\begin{equation}
h(\mbQ_t, \mbK_t) = \frac{\mbQ_t \mbK_t^\top + \lambda \mbM_t^{\text{fair}}}{\sqrt{d}}.
\end{equation}

The fairness-aware modulation term $\mbM_t^{\text{fair}} \in \mathbb{R}^{(H \cdot W) \times L}$ is constructed as:
\begin{equation}
\mbM_t^{\text{fair}} = \mbR_t \odot \mbM_t^{\text{pos}} - (1 - \mbR_t) \odot \mbM_t^{\text{neg}},
\end{equation}
with
\begin{equation}
\begin{aligned}
\mbM_t^{\text{pos}} &= \max(\mbQ_t \mbK_t^\top) - \mbQ_t \mbK_t^\top, \\
\mbM_t^{\text{neg}} &= \mbQ_t \mbK_t^\top - \min(\mbQ_t \mbK_t^\top).
\end{aligned}
\end{equation}
and the soft similarity mask $\mbR_t[\texttt{q}, \texttt{k}]$ is defined by
\begin{equation}
\mbR_t[\texttt{q}, \texttt{k}] =
\begin{cases}
\cos\!\big(\mbQ_t[\texttt{q}], \mbe_k\big), & \text{if } \texttt{k} \in \tau_k, \\
0, & \text{otherwise}.
\end{cases}
\end{equation}
 Here, $\texttt{q} \in \{1, \dots, H \cdot W\}$ indexes the spatial query positions, and $\texttt{k} \in \{1, \dots, L\}$ indexes the prompt token positions. $\mbe_k \in \mathbb{R}^d$ denotes the embedding of the $k$-th fairness token, and $\tau_k$ is the set of key indices corresponding to $\mbe_k$. The cosine similarity softly aligns each query to its semantically relevant fairness concept, allowing fine-grained attention modulation. This soft similarity mask $\mbR_t$ biases the attention logits toward fairness-relevant tokens, ensuring that spatial queries focus more on semantically aligned fairness concepts while suppressing unrelated prompts. Unlike a hard mask, the cosine similarity provides a continuous modulation, preserving the natural attention distribution while improving fine-grained prompt-to-region alignment.

This approach dynamically guides attention toward debiasing-relevant tokens while mitigating interference from irrelevant prompts, thereby enhancing fairness-aware consistency across frames in video generation.

% We define a cross-attention operator with fairness-guided modulation as:
% \begin{equation}
% \texttt{FairCrossAttention}(\mbQ, \mbK, \mbV) = \mathrm{softmax}\left(h(\mbQ, \mbK)\right) \cdot \mbV,
% \end{equation}
% where the score function $h(\mbQ, \mbK)$ is defined as:
% \begin{equation}
% h(\mbQ, \mbK) = \frac{\mbQ \mbK^\top + \lambda \mbM^{\text{cross}}}{\sqrt{d}}.
% \end{equation}

% The modulation term $\mbM^{\text{cross}}$ is constructed as:
% \begin{equation}
% \mbM^{\text{cross}} = \mbR \odot \mbM^{\text{pos}} - (1 - \mbR) \odot \mbM^{\text{neg}},
% \end{equation}
% where:
% \begin{equation}
% \begin{aligned}
% \mbM^{\text{pos}} &= \max(\mbQ \mbK^\top) - \mbQ \mbK^\top, \\
% \mbM^{\text{neg}} &= \mbQ \mbK^\top - \min(\mbQ \mbK^\top),
% \end{aligned}
% \end{equation}
% and the region mask $\mbR$ is defined based on fairness token positions:
% \begin{equation}
% \mbR[x, y] =
% \begin{cases}
% m_{i,k}, & \text{if } y \in \tau_k, \\
% 0, & \text{otherwise},
% \end{cases}
% \end{equation}
% where $\tau_k$ indexes the fairness-relevant tokens, and $m_{i,k} \in [0,1]$ denotes the soft spatial relevance of pixel $x$ to concept $e_k$.

% This design softly increases attention weights between fairness-related prompts and their corresponding spatial regions, while suppressing attention to unrelated areas. It allows attribute-guided alignment to be propagated across frames in a smooth and controllable manner.

\section{Experiments}
\subsection{FairVE: Datasets for Fairness in Frame-Level Video Editing}

\textbf{Data Collection and Fairness Configurations.} Following previous studies~\cite{zhang_camel_2024,jeong_dreammotion_2024}, we construct the \textit{FairVE} dataset consisting of video–text pairs sourced from the DAVIS benchmark and supplemented with high-quality, user-generated videos collected from the web. Rather than relying on artificially designed samples, FairVE reflects naturally occurring social bias in real-world data, ensuring realistic evaluation of fairness-aware VE methods. To better capture and evaluate social bias, we restrict the dataset to scenes containing a single human subject, filtering out clips with animals, vehicles, multiple individuals, or complex backgrounds. The corresponding text prompts are automatically generated using GPT-4~\cite{leonardis_wave_2025} to ensure consistency across samples. Each video sequence contains 30 to 60 frames with $512 \times 512$ or $640 \times 320$ pixels. 

In the fairness evaluation protocol, we focus on gender-profession stereotypes observed in SD 1.5~\cite{kim_rethinking_2025}, as subclass of ~\cite{orgad2023editing,chuang2023debiasing}. Eight representative professions are selected: CEO, doctor, technician, pilot, teacher, nurse, fashion designer and librarian. Specifically, we combine each profession with gender-specific terms (``male” or ``female”) to create 640 prompts (8 professions $\times$ 2 genders $\times$ 40 videos).

\textbf{Fairness Evaluation}~We assess the bias correction capability followed by ~\cite{kim_rethinking_2025} and complement it with human evaluations. For each profession specific prompt, FAME is repeated 5 times under different random seeds. 

For automatic metrics, We record two metrics: (1) \textit{Correction Count}, the total number of trials in which the edited result shows reduced gender bias compared with the unedited baseline; and (2) \textit{Correction Ratio}, the relative frequency of such bias-reducing outcomes. Higher values indicate more consistent mitigation.  

To provide a qualitative perspective, we conduct human evaluation~\cite{friedrich_fair_2023,li2025fair,li2025t2isafety} along three dimensions: \textit{Perceived Bias Correction}, whether the generated portrayal of the profession appears unbiased; \textit{Visual Stability}, the consistency of appearance and motion across frames; and \textit{Semantic Relevance}, the alignment between generated content and the intended meaning of the prompt. Twenty participants rated 640 video-text pairs on a scale from 20 to 100 for each aspect. Averaged scores serve as complementary evidence to the statistical results.

\textbf{Baselines}~We compare our model’s fidelity, consistency, and fairness with several SOTA training-free VE methods, including FateZero~\cite{qi_fatezero_2023}, TokenFlow~\cite{geyer2023tokenflow}, DMT~\cite{yatim2024space}, Slicedit~\cite{cohen_slicedit_nodate} and VideoGrain~\cite{yang_videograin_2025}. Since these baselines do not incorporate fairness mechanisms by design, we adopt a prompt-based mitigation strategy by explicitly injecting bias debiasing terms into their input prompts. We then evaluate both their VE quality and fairness performance using the same metrics, allowing a direct comparison with FAME.

\begin{table*}[ht]
  \centering
  \small
  \begin{tabular}{cc||cc|cccc}
    \toprule

    \multirow{2}{*}{\textbf{Minor Group}} & 
    \multirow{2}{*}{\textbf{Professions}} & 
    \multicolumn{2}{c|}{\textbf{Automatic}} &
    \multicolumn{4}{c}{\textbf{User Study}} \\
    \cmidrule(lr){3-4} \cmidrule(lr){5-8}
     &&\textbf{Count} & \textbf{Ratio} & \textbf{Bias Correction}$\uparrow$ & \textbf{Visual Stability}$\uparrow$ & \textbf{Semantic Relevance}$\uparrow$ & \textbf{Overall}$\uparrow$ \\
    \midrule
    \multirow{4}{*}{\rotatebox{0}{Female}} 
    & CEO &9 & 0.45 & 70.2 & 40.2 & 86.7 & 65.70\\
    & Doctor &6 & 0.30 & 83.7& 44.8 & 89.2&72.57\\
    & Technician &4 & 0.20 & 80.8& 55.4& 88.5&74.90\\
    & Pilot &10 & 0.50 & 84.6& 48.3& 90.1&74.33\\
    \midrule
    \multirow{4}{*}{\rotatebox{0}{Male}} 
    & Teacher &10 & 0.50 & 69.0& 59.0& 79.6&69.20\\
    & Nurse &2 & 0.10 &88.2 & 60.8& 91.1& 80.03\\
    & Fashion designer &9 & 0.45 &71.8 & 53.1 &86.9 &70.60\\
    & Librarian &9 & 0.45 & 72.9& 59.2& 87.4&49.83\\
    \bottomrule
  \end{tabular}
  
  \caption{Fairness evaluation across professions and minority groups.}
  \label{tab:fair}
\end{table*}

\begin{table}[ht]
\centering
\scriptsize
% \resizebox{0.51\textwidth}{!}{
\begin{tabular}{c||ccc|c}
\toprule
& \multicolumn{3}{c|}{\textbf{CLIP-based}} & \textbf{Other} \\
\textbf{Method}  & \textbf{CLIP-F} $\uparrow$ & \textbf{CLIP-T} $\uparrow$ & \textbf{TIFA} $\uparrow$ & \textbf{Warp-Err} $\downarrow$ \\
\midrule
FateZero & 95.75 & 33.78 & 48.65 &  3.08   \\
TokenFlow  & 96.48 & 34.59 &  49.62 & 2.82 \\
Ground-A-Video &95.17  &35.09  & / & 4.43 \\
DMT & 96.34 & 34.09 & / & 2.05 \\
VideoGrain &  97.18& 35.22  & 60.17 & 2.21  \\
\midrule
% \rowcolor[gray]{0.9}
\textbf{FAME (Ours)} & \textbf{98.31} & \textbf{35.9} & \textbf{62.2} & \textbf{1.66} \\
\bottomrule
\end{tabular}
% }
\caption{Quantitative comparison of automatic metrics with FateZero~\cite{qi_fatezero_2023}, TokenFlow~\cite{geyer2023tokenflow}, Ground-A-Video~\cite{jeong_ground--video_2024}, DMT~\cite{yatim2024space} ,VideoGrain~\cite{yang_videograin_2025}.}
\label{table:VE}
\end{table}

\subsection{Experimental Results}

\textbf{Fairness Comparison}~Fairness results are shown in \Cref{tab:fair}. From the automatic metrics, professions with strong stereotypes such as \textit{Doctor} and \textit{Nurse} show clear improvements, with high correction ratios and strong bias correction scores. Human evaluation further confirms this: \textit{Bias Correction} ratings exceed 80/100, indicating effective mitigation, while \textit{Semantic Relevance} remains high, reflecting faithful prompt alignment. \textit{Visual Stability} scores are relatively lower, suggesting temporal consistency still needs improvement. Overall, these results show that our method reduces bias effectively while maintaining semantic fidelity, though motion continuity can be further enhanced.

\textbf{Video Editing Comparison}~As shown in ~\Cref{table:VE}, FAME outperforms all baselines across key CLIP-based metrics, including CLIP-F, CLIP-T~\cite{cong2023flatten}, and TIFA~\cite{hu2023tifa}, demonstrating visual-text alignment, temporal coherence, and image-text fidelity, respectively. Comparing with the results in ~\cite{yang_videograin_2025,fan_vs_nodate}, we re-run the metrics for VideoGrain as a baseline (results are taken directly from~\cite{yang_videograin_2025} where our experimental dataset inherits and the comparison is fair). FAME also achieves the lowest warp error, demonstrating better temporal consistency and spatial coherence in the edited sequences. Unlike previous methods, which rely on latent inversion or token warping, FAME incorporates attention modulation, leading to more visually stable and semantically accurate results, especially under attribute-altering prompts. For more visualization results, please refer to our GitHub Pages. The anonymous repository link is available in Appendix~\ref{sec:github}.

\begin{figure}[t]
  \centering
  \includegraphics[width=\linewidth]{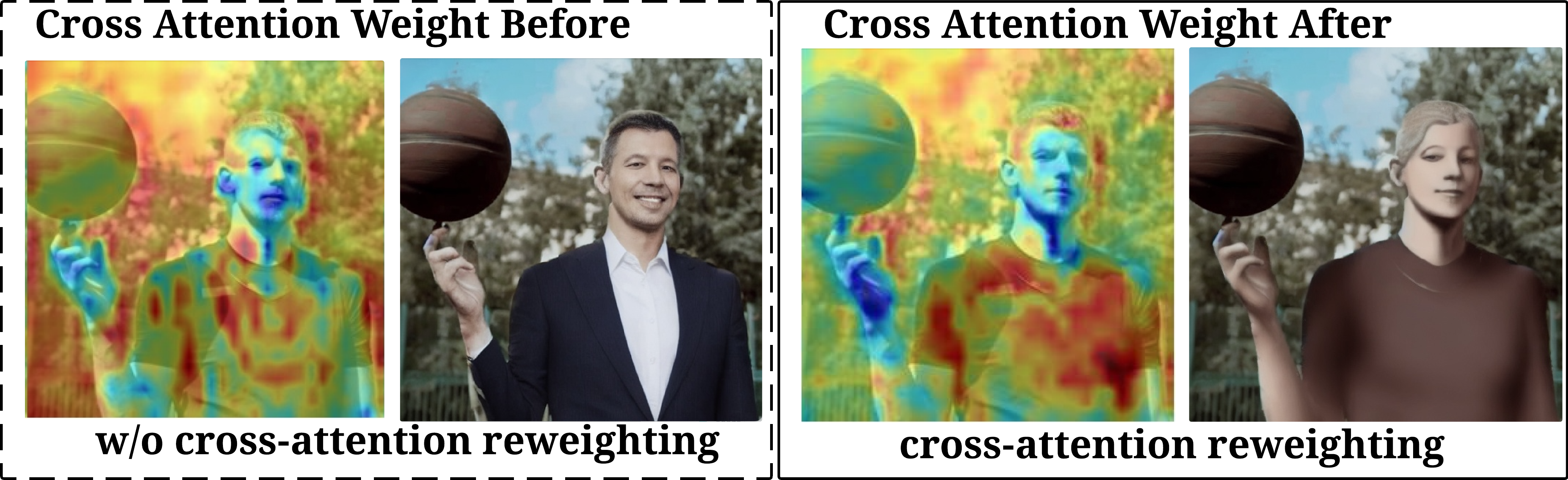}
  \caption{
  \textbf{Cross-Attention Modulation Effect.} Visualization of attention maps corresponding to the token $\texttt{CEO}$ under different ablation settings. The baseline model struggles to separate foreground(human) and background(tree), often overlooking regions related to fairness. Cross-attention reweighting helps adjust areas like outfit to better preserve identity features.
  }
  \label{fig:fair_cross_attn}
\end{figure}

\subsection{Ablation Study}
\textbf{Cross-Attention Disentanglement} 
To evaluate whether our attention modulation enhances fairness-relevant region separation, we design a simplified version of FAME. Each frame is segmented into a foreground (e.g., man) and a background (e.g., trees) region. We use the prompt \textit{``A CEO is spining ball"} to examine cross-attention responses. Without our modulation, the attention weights for the word ``CEO'' tend to spread into the background, causing semantic interference. As a result, the edited region retains biased visual features, preserving the male outfit. In contrast, with cross-attention reweighting, attention weights become more localized, emphasizing the foreground region while suppressing irrelevant background responses. In that case, the foreground region will be modified more to debiasing content(the outfit of female). This separation is essential for fairness-aware editing, ensuring that debiasing prompt related to protected concepts remain spatially and semantically disentangled from unrelated content. As shown in~\Cref{fig:fair_cross_attn}, our module produces a clearer attention focus aligned with the intended semantic region. 
% Additional visual comparisons are provided in the anonymous GitHub pages.

\subsection{Attention Modulation for Fairness Control}

We conduct an ablation study to verify the effectiveness of our fairness-guided attention modulation. Specifically, we compare three variants: 
(1) \textbf{Baseline}, which applies text to VE with prompt-guided module; 
(2) \textbf{+ Self-Attn Modulation}, where we introduce fairness alignment using our modulation map $R_t$; and 
(3) \textbf{+ Self-Attn + Cross-Attn Modulation}, the full version of our framework.

% As shown in Fig.~\ref{fig:fair_cross_attn}, the editing prompt is \textit{``A firefighter is playing tennis''}. Without cross-attention modulation, the model fails to localize fairness-related tokens, resulting in entangled visual edits or identity mismatch. In contrast, with our proposed modulation, the attention on fairness-critical tokens such as ``firefighter'' is sharply aligned with the intended visual region across frames, producing faithful and consistent edits.

Quantitative results in~\Cref{table:ab} show that adding self-attention and cross-attention modulation progressively improves CLIP-based metrics, with CLIP-F increasing slightly to $98.30$, CLIP-T to $35.80$, and TIFA to $61.22$, compared to the baseline. Meanwhile, the fairness ratio of male teacher maintain in $0.50$, indicating the fairness can be achieved combined with three modules.

\subsection{Sensitivity Analysis on Text Encoding}

To investigate the influence of our fairness-guided prompt encoding strategy, we conduct two controlled sensitivity 
analyses focusing on the \textit{modulation strength} of the debiasing embedding.
% analyses focusing on (1) the \textit{injection method} and (2) the \textit{modulation strength} of the debiasing embedding.

% \textbf{Injection Method Comparison.}  
% We compare two strategies for integrating the bias-type embedding $\mbe_{\texttt{fair}}$ into the prompt embedding: (a) \textit{Direct Addition}, where $\mbp_{\texttt{fair}}$ is uniformly added to all token embeddings $\mbe_{\texttt{tar}}$, and (b) \textit{Masked Addition} (our method), where $\mbp_{\texttt{fair}}$ is fused into semantically relevant positions via a binary mask $M$. As shown in Appendix~\Cref{fig:prompt-test}, direct injection results in semantic drift in lines 2,3 and loss of debiasing property specificity in line 4. Here, we compare the FAME with other baselines.

\textbf{Modulation Strength Sensitivity.}  
We further study the impact of varying the injection strength by introducing a scaling factor $\alpha$:
\begin{equation}
g(\mbp_{\texttt{fair}}, \mbp_{\texttt{tar}}) = \tilde{\mbe}_{\texttt{tar}} + \alpha \mathbf{M} \otimes (\mbe_{\texttt{fair}} - \tilde{\mbe}_{\texttt{tar}}),
\end{equation}
where $\alpha \in \{0.0, 0.25, 0.5, 1.0\}$. 

As illustrated in Appendix ~\Cref{fig:text_injection}, smaller values of $\alpha$ result in underwhelming fairness control (e.g., no visible change in target appearance), while excessively large $\alpha$ may distort original semantics. Our method achieves the best trade-off when $\alpha = 0.5$, offering both fairness alignment and semantic stability.

\begin{table}[ht]
\centering
\scriptsize
% \resizebox{0.49\textwidth}{!}{
\begin{tabular}{c||ccc|c}
\toprule
\multirow{2}{*}{\textbf{Ablation Settings}} 
& \multicolumn{3}{c|}{\textbf{VE Quality}} & \textbf{Fairness} \\
\cmidrule(lr){2-4} \cmidrule(lr){5-5}
& \textbf{CLIP-F} $\uparrow$ & \textbf{CLIP-T} $\uparrow$ & \textbf{TIFA} $\uparrow$ & \textbf{Ratio}  \\
\midrule
\texttt{+P} & 95.70 & 33.99 & 49.6 & / \\
\texttt{+P +S} & 98.10 & 35.10 & 60.37 & 0.50 \\
% \rowcolor[gray]{0.9}
\texttt{+P +S +C} & \textbf{98.30} & \textbf{35.80} & \textbf{61.22} & \textbf{0.50} \\
\bottomrule
\end{tabular}
% }
\caption{
Ablation study of key modules. 
\texttt{P}: \textit{Prompt-guided module}; 
\texttt{S}: \textit{Self-attention module}; 
\texttt{C}: \textit{Cross-attention module}. 
Modules are added cumulatively from top to bottom.
}
\label{table:ab}
\end{table}

\section{Conclusion}

In this paper, we present a training-free framework FAME for fairness-aware VE that addresses the limitations of existing diffusion-based models in handling fairness-sensitive prompts. By introducing soft debiasing prompt encoding and fairness-guided modulation in both self-attention and cross-attention layers, FAME ensures improved alignment between textual attributes and visual regions while preserving temporal coherence. To support systematic evaluation, we construct FairVE, the first benchmark dataset tailored for assessing fairness in VE tasks, encompassing diverse gender–profession combinations. Extensive experiments, including automatic metrics and structured human evaluations, show that FAME outperforms state-of-the-art zero-shot baselines in semantic fidelity, visual consistency, and bias mitigation. Because FAME works entirely at the inference time, it avoids expensive retraining yet still improves fairness and stability. We will extend this strategy for broader fairness-aware VE research.

\clearpage
\newpage

\bibliography{aaai26}

% \clearpage
% \input{ReproducibilityChecklist.tex}

\clearpage
\newpage
\section{Appendix}

% \subsection{Github Pages of FAME}
% \label{sec:github}

% We have publish our project and code on anonymous link. For more visulization results (including the Prompt Responsiveness Test and editing results) and experimental details please refer to: https://anonymous.4open.science/w/FAME-0821/

\subsection{Existing literature on Fairness in T2I methods}

Prior work on fair T2I mostly falls into three camps: adding semantic guidance during diffusion optimization~\cite{dewan2024diffusion, shen2023finetuning, jung2025multi}, injecting debiasing embeddings into the latent space~\cite{huang2025implicit, azam2025plug, li2025responsible},, or applying classifier-guided noise control\cite{brack2023sega, kim_rethinking_2025}. Lightweight alternatives modify attention maps~\cite{gandikota_unified_2024, teo2024fairqueue, zhou2024association, park2025fair} or adjust the bottleneck representations in the U-Net~\cite{li2024self, shi2025dissecting}, avoiding full model retraining.

\subsection{Implementation Details}
\textbf{VE Evaluation Metrics}~To demonstrate that our model effectively preserves VE capabilities, we adopt a suite of established CLIP similarity metrics~\cite{radford2021learning} that capture both edit fidelity and temporal consistency. We primarily focus on CLIP-based metrics to measure semantic alignment and structural coherence. Specifically, we employ CLIP-T, which calculates the average cosine similarity between the input text prompt and all generated video frames using CLIP embeddings, thereby reflecting how well the output video aligns with the intended prompt semantics. CLIP-F measures the average cosine similarity between consecutive video frames, serving as an indicator of temporal coherence within the generated sequence. TIFA score evaluates semantic alignment between the target image and the edited video frames. Additionally, Warp-Err quantifies the pixel-level deviation by warping each edited frame based on the optical flow of the source video using \textit{RAFT-Large}~\cite{teed2020raft}, thereby evaluating motion consistency. 

\textbf{Experimental Settings}~We build our framework upon SD 1.5\footnote{https://huggingface.co/stable-diffusion-v1-5/stable-diffusion-v1-5}, utilizing its released pre-trained weights along with the DDIM scheduler~\cite{song2020denoising} with 50 steps for inverse to acquarie attention maps. Each video–text pair is represented as the tuple \textnormal{(source video, instruction, SAM masks, ControlNet annotators)}. We employ SAM-tracking~\cite{cheng2023segment} based masks to perform object-level clustering, enabling fine-grained edge segmentation. To support controllable image-level editing, we integrate several ControlNet annotators~\cite{zhang2023adding}, including DW-Pose, Depth-Zoe, Depth-MiDaS, and OpenPose (totaling approximately 4 GB). We retain the default \texttt{guidance\_scale} parameter to control the influence of the prompt on the editing outcome. Additionally, we enable Prompt-to-Prompt (PnP)~\cite{hertz2022prompt} editing to modulate spatial or semantic regions across frames. All experiments are conducted on a single NVIDIA Tesla H100 GPU.
\subsection{Additional Results}
\begin{figure}[t]
  \centering
  \includegraphics[width=\linewidth]{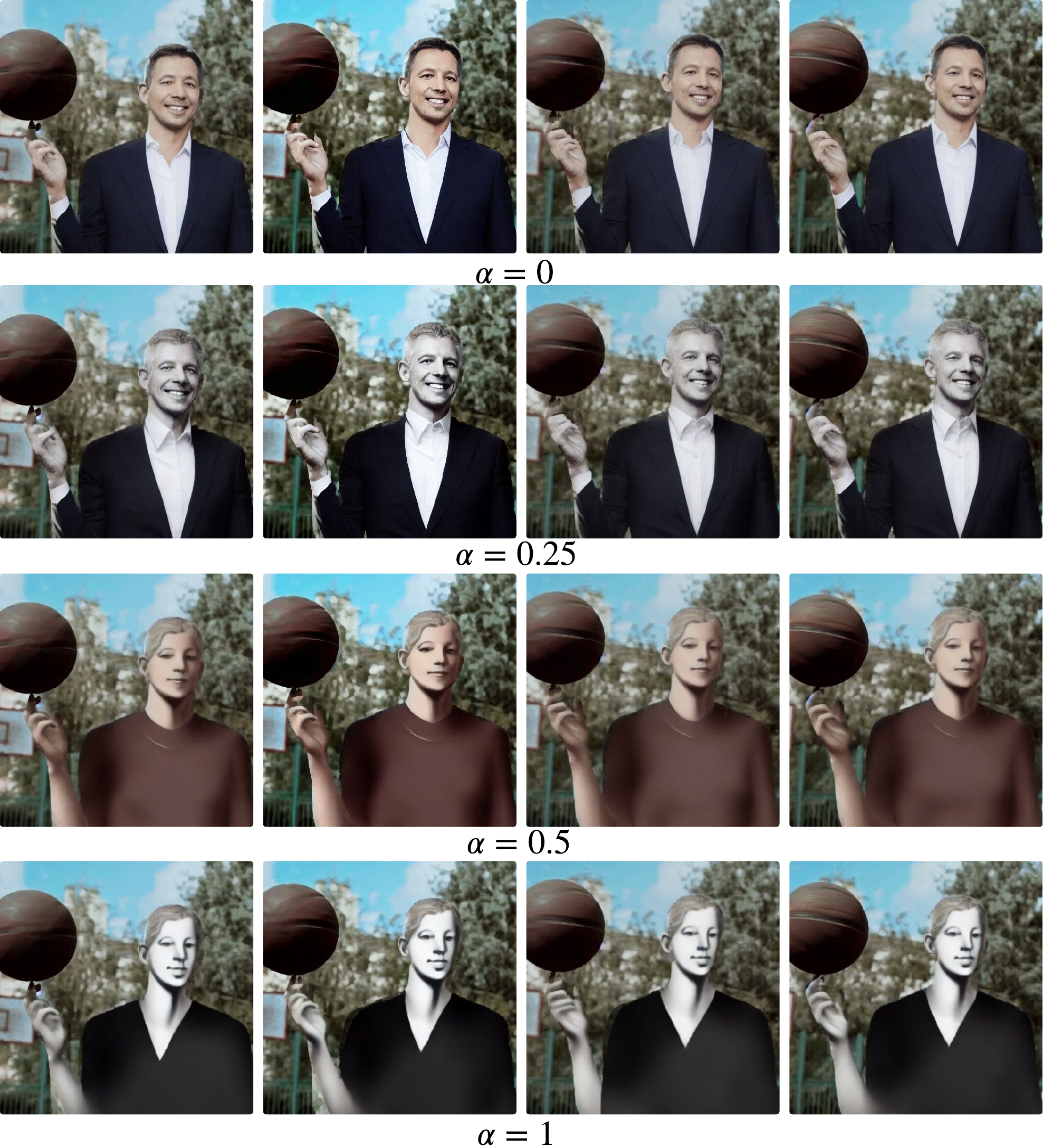}
  \caption{Comparison of different prompt injection methods. Direct addition leads to semantic confusion, while our masked fusion retains structure and enhances fairness controllability.}
  \label{fig:text_injection}
\end{figure}

\Cref{fig:text_injection} verify the necessity of the importance of balanced modulation strength for fairness-aware prompt design in T2I based training-free VE models.

\begin{figure*}
  \centering
  \includegraphics[width=1\linewidth]{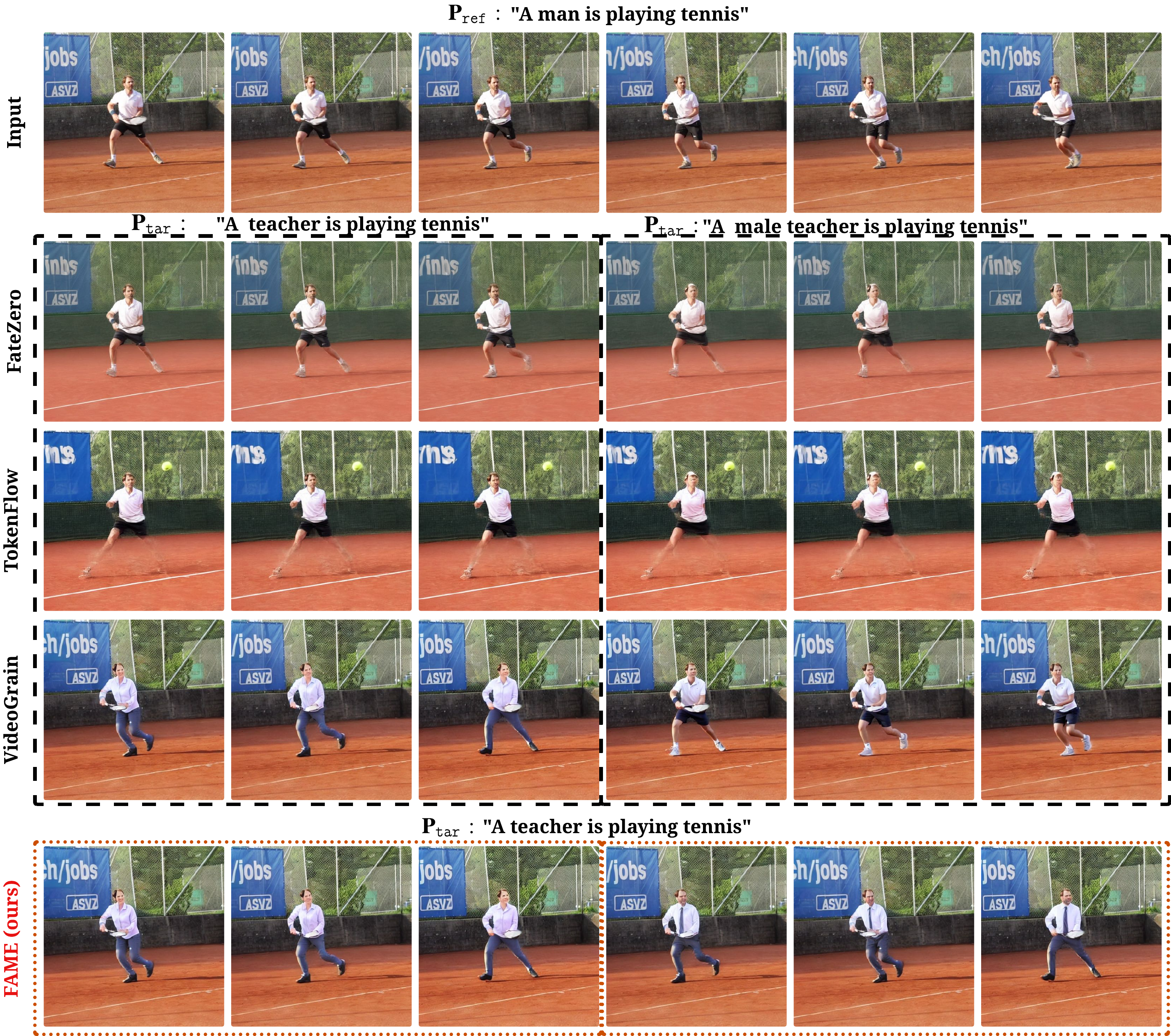}
  \caption{
  \textbf{Prompt Responsiveness Test.} We evaluate three training-free VE models, including FateZero, TokenFlow, and VideoGrain, on a single example $\mbP_{\texttt{ref}}=\texttt{"A man is playing tennis"}$ to assess their responsiveness to debiasing prompt. The top row shows the input reference frames. Rows 2–4 display results using different prompts on several baselines, including FateZero~\cite{qi_fatezero_2023}, TokenFlow~\cite{geyer2023tokenflow}, and VideoGrain~\cite{yang_videograin_2025}, while the 5th row shows the results with the original editing prompt $\mbP_{\texttt{tar}}=\texttt{"A teacher is playing tennis"}$ by our FAME. In rows 2–4, the left column shows the editing performance with the original prompt, and the right column shows the editing performance guided by the directly injected prompt $\mbP_{\texttt{tar}}=\texttt{"A male teacher is playing tennis"}$ as the debiasing prompt. In the last row, our FAME can generate the female version (left column) and the debiased male content (right column). In the 2nd and 3rd rows, FateZero and TokenFlow exhibit a mismatch between the debiasing prompt and the visual semantics, indicating degraded video fidelity. In the 4th row, VideoGrain shows that directly injecting debiasing prompts often leads to semantic degradation in generated frames. This suggests that fairness-sensitive prompts may disrupt the alignment between the prompt and visual output. In the last row, our FAME can generate different gender-profession content even with the original prompt.
  }
  \label{fig:prompt-test}
\end{figure*}

\Cref{fig:prompt-test} shows the full test pipeline and results for the  Prompt Responsiveness Test mentioned in \Cref{sec:method}. 

\textbf{Pseudo Algorithm Code.} 
Our full inference procedure is composed of three core modules, shown in \Cref{alg:fair-embed},\Cref{alg:fair-temporal-attn} and \Cref{alg:fair-cross-attn}. \Cref{alg:fair-embed} describes the fairness-aware text embedding procedure, which softly fuses debiasing attribute tokens into the original prompt embedding to retain semantic structure while introducing fairness control. \Cref{alg:fair-temporal-attn} defines the fairness-aware self-attention modulation that enhances intra-region coherence by amplifying attention within fairness-relevant areas and suppressing unrelated interactions. \Cref{alg:fair-cross-attn} introduces the fairness-aware cross-attention mechanism, where alignment between prompt tokens and visual regions is refined using a similarity-based attention mask.

\begin{algorithm}[!htbp]
\caption{Soft Debiasing Prompt Encoding via Position-Aware Fusion}
\label{alg:fair-embed}
\begin{algorithmic}[1]
\REQUIRE Target prompt $\mbp_{\texttt{tar}}$, fairness prompt $\mbp_{\texttt{fair}}$, pre-trained text encoder $\mbphi$
\ENSURE Debiased embedding matrix $\tilde{\mbe} \in \mathbb{R}^{L \times d}$

\STATE $\mbe_{\texttt{tar}} \gets \mbphi(\mbp_{\texttt{tar}})$ \hfill // Token embeddings of target prompt
\STATE $\mbe_{\texttt{fair}} \gets \mbphi(\mbp_{\texttt{fair}})$ \hfill // Token embeddings of fairness prompt
\STATE $\mathbf{M} \gets \textsc{GenerateSoftMask}(\mbe_{\texttt{tar}}, \mbe_{\texttt{fair}})$ \hfill // Soft mask: $\mathbf{M} \in [0,1]^{L \times d}$
\STATE $\tilde{\mbe} \gets \mbe_{\texttt{tar}} + \mathbf{M} \otimes (\mbe_{\texttt{fair}} - \mbe_{\texttt{tar}})$ \hfill // Soft injection
\RETURN $\tilde{\mbe}$
\end{algorithmic}
\end{algorithm}

\begin{algorithm}[!htbp]
\caption{Self-attention Manipulation}
\label{alg:fair-temporal-attn}
\begin{algorithmic}[1]
\REQUIRE Latent tensor $\mbz_t \in \mathbb{R}^{h \times w \times \ell \times c}$, projections $W_\mbQ, W_\mbK, W_\mbV$, coefficients $\lambda$, $\mbmu$, temperature $\tau$
\ENSURE Attention output $\mbA \in \mathbb{R}^{(h \cdot w) \times \ell \times d}$

\STATE $\mbz' \gets \varphi(\mbz_t)$ \hfill // Flatten spatial dims to $(h \cdot w) \times \ell \times c$
\STATE $\mbQ, \mbK, \mbV \gets W_\mbQ \mbz',\ W_\mbK \mbz',\ W_\mbV \mbz'$ \hfill // Project to attention space
\STATE $\mathbf{A}_{\text{raw}} \gets \mbQ \mbK^\top$ \hfill // Raw attention scores
\STATE Compute $\mbM^{\text{pos}}, \mbM^{\text{neg}}$ from $\mathbf{A}_{\text{raw}}$
\STATE Compute region mask $\mbR$ \hfill // 1 if same region, else 0
\STATE $\mbM^{\text{fair}} \gets \mbR \odot \mbM^{\text{pos}} - (1 - \mbR) \odot \mbM^{\text{neg}}$
\STATE Compute per-position feature mean: $\mathbf{f}_{\texttt{q}} = \frac{1}{\ell} \sum_{t=1}^{\ell} \mbz_t[\texttt{q}]$
\STATE Compute similarity mask: $\mbS[\texttt{q}_1, \texttt{q}_2] = \exp(-\|\mathbf{f}_{\texttt{q}_1} - \mathbf{f}_{\texttt{q}_2}\|^2 / \tau^2)$
\STATE $g(\mbQ, \mbK) \gets (\mathbf{A}_{\text{raw}} + \lambda \mbM^{\text{fair}} + \mbmu \mbS)/\sqrt{d}$
\STATE $\mbA \gets \text{softmax}(g(\mbQ, \mbK)) \cdot \mbV$ \hfill // Final output
\RETURN $\mbA$
\end{algorithmic}
\end{algorithm}

\begin{algorithm}[!htbp]
\caption{Cross-attention Reweighting}
\label{alg:fair-cross-attn}
\begin{algorithmic}[1]
\REQUIRE $\mbQ_t, \mbK_t, \mbV_t$, fairness token set $\{\mbe_k\}$, token groups $\{\tau_k\}$, coefficient $\lambda$
\ENSURE Cross-attention output $\mbA_t$

\STATE $\mathbf{A}_{\text{raw}} \gets \mbQ_t \mbK_t^\top$ \hfill // Raw attention matrix
\STATE Compute $\mbM_t^{\text{pos}}, \mbM_t^{\text{neg}}$ from $\mathbf{A}_{\text{raw}}$
\FOR{each spatial query index $\texttt{q}$}
    \FOR{each prompt token index $\texttt{k}$}
        \IF{$\texttt{k} \in \tau_k$}
            \STATE $\mbR_t[\texttt{q}, \texttt{k}] \gets \cos(\mbQ_t[\texttt{q}], \mbe_k)$ \hfill // Fairness match
        \ELSE
            \STATE $\mbR_t[\texttt{q}, \texttt{k}] \gets 0$
        \ENDIF
    \ENDFOR
\ENDFOR
\STATE $\mbM_t^{\text{fair}} \gets \mbR_t \odot \mbM_t^{\text{pos}} - (1 - \mbR_t) \odot \mbM_t^{\text{neg}}$
\STATE $h(\mbQ_t, \mbK_t) \gets (\mathbf{A}_{\text{raw}} + \lambda \mbM_t^{\text{fair}})/\sqrt{d}$
\STATE $\mbA_t \gets \text{softmax}(h(\mbQ_t, \mbK_t)) \cdot \mbV_t$ \hfill // Cross-attention output
\RETURN $\mbA_t$
\end{algorithmic}
\end{algorithm}

\end{document}

%% file: math.tex
\newcommand{\mathbold}[1]{\ensuremath{\boldsymbol{\mathbf{#1}}}}

\newcommand{\nestedmathbold}[1]{{\mathbold{#1}}}

\newcommand{\mba}{\nestedmathbold{a}}

\newcommand{\mbe}{\nestedmathbold{e}}

\newcommand{\mbp}{\nestedmathbold{p}}

\newcommand{\mbx}{\nestedmathbold{x}}

\newcommand{\mbz}{\nestedmathbold{z}}

\newcommand{\mbA}{\nestedmathbold{A}}

\newcommand{\mbI}{\nestedmathbold{I}}

\newcommand{\mbK}{\nestedmathbold{K}}

\newcommand{\mbM}{\nestedmathbold{M}}

\newcommand{\mbP}{\nestedmathbold{P}}
\newcommand{\mbQ}{\nestedmathbold{Q}}
\newcommand{\mbR}{\nestedmathbold{R}}
\newcommand{\mbS}{\nestedmathbold{S}}

\newcommand{\mbV}{\nestedmathbold{V}}

\newcommand{\mbX}{\nestedmathbold{X}}

\newcommand{\mbepsilon}{\nestedmathbold{\epsilon}}

\newcommand{\mbmu}{\nestedmathbold{\mu}}

\newcommand{\mbphi}{\nestedmathbold{\phi}}

\newcommand{\mbtheta}{\nestedmathbold{\theta}}